\pdfoutput=1

\documentclass[11pt]{article}

\usepackage[preprint]{acl}

\usepackage{times}
\usepackage{latexsym}

\usepackage[T1]{fontenc}

\usepackage[utf8]{inputenc}

\usepackage{microtype}

\usepackage{inconsolata}

\usepackage{graphicx}

\usepackage{times}
\usepackage{url}
\usepackage{latexsym}
\usepackage{booktabs}
\usepackage{multirow}
\usepackage{adjustbox}
\usepackage{hyperref}
\usepackage{graphicx}
\usepackage{tikz-dependency}

\title{The Language of Attachment: \\ Modeling Attachment Dynamics in Psychotherapy}

\author{
Frederik Bredgaard\textsuperscript{1}\hspace{1em}
Martin Lund Trinhammer\textsuperscript{2,3}\hspace{1em} 
Elisa Bassignana\textsuperscript{2,3}\hspace{1em} \\
\textsuperscript{1}SODAS, University of Copenhagen, Denmark \hspace{1em} \\
\textsuperscript{2}IT University of Copenhagen, Denmark \hspace{1em} \\
\textsuperscript{3}Pioneer Center for Artificial Intelligence, Denmark \\
{\tt frederik@bredgaard.dk}\hspace{.9em}{\tt \{mlut,elba\}@itu.dk}
  }

\begin{document}
\maketitle
\begin{abstract}
    The delivery of mental healthcare through psychotherapy stands to benefit immensely from developments within Natural Language Processing (NLP), in particular through the automatic identification of patient specific qualities, such as attachment style.
    Currently, the assessment of attachment style is performed manually using the Patient Attachment Coding System (PACS; \citeauthor{Talia2017}, \citeyear{Talia2017}), which is complex, resource-consuming and requires extensive training. To enable wide and scalable adoption of attachment informed treatment and research, we propose the first exploratory analysis into automatically assessing patient attachment style from psychotherapy transcripts using NLP classification models. 
    We further analyze the results and discuss the implications of using automated tools for this purpose---e.g., confusing `preoccupied' patients with `avoidant' likely has a more negative impact on therapy outcomes with respect to other mislabeling.
    Our work opens an avenue of research enabling more personalized psychotherapy and more targeted research into the mechanisms of psychotherapy through advancements in NLP.
\end{abstract}

\section{Introduction}
As an exclusively language-driven type of therapy for mental illnesses, psychotherapy may be revolutionized through application of methods from Natural Language Processing (NLP) over the course of the coming decade \cite{stade2024large}. Application scenarios are manifold as fundamental questions underpinning psychotherapy remain unanswered. For instance, it is clear that patients benefit significantly from psychotherapy, however, the mechanism through which psychotherapy works is largely a black box \cite{cuijpers2019role}. 
Given the highly rich nature of a patient's vocalized introspection, multiple efforts within psychotherapy research have been taken to extract meaningful constructs from the therapeutic encounter. If such measures are validly and reliably identified, they could function as future treatment objectives, giving the therapists coherent guidelines to target, rather than relying on intuition or own past experience. 
One such tool, the PACS \cite{Talia2017}, can reliably infer patients attachment style only by analyzing the speaking pattern of the patient. The theory of attachment remains one of the most robust and extensively validated theories in psychology \cite{Cassidy2013, Slade2016, Gregory2020}. The theory describes how the relationship between a child and its caregiver forms an affectional tie which endures throughout the child's life \cite{Bowlby1988, Ainsworth1970}. The child's attachment style eventually comes to influence almost all relations, ranging from close family and romantic partners to friends and professional relationships \cite{Rosenthal2010, Karantzas2011ArthritisAS, Rowe2005, Tancredy2006}. 
Much research underscores the significant impact attachment-informed psychotherapy holds for long-term symptom alleviation and recovery of patients \cite{Shorey2006, Slade2019}. A significant bottleneck to scaling this kind of psychotherapy is the fact that existing manual methods (e.g. PACS, \citeauthor{Talia2017}, \citeyear{Talia2017}; AAI, \citeauthor{AAIProtocol}, \citeyear{AAIProtocol}) are time-consuming to administer and require highly trained expert annotators, rendering practical use unfeasible.

In this paper, we employ NLP technologies to investigate the automatic classification of patient-attachment from psychotherapy transcripts. 
We perform turn-level classification and employ the RoBERTa-large encoder~\cite{roberta}, as well as the domain-specific MentalRoBERTa~\cite{MentalBERT}. We combine both architectures with domain adaptive pre-training on unlabeled data~\cite{Gururangan2020}. Our best results achieve an average accuracy of almost 60\% in distinguishing the three available attachment classes.
Given the nature of psychotherapy, there is high variability between the length and the amount of information carried in each patient turn. Therefore, we further investigate these aspects with a series of experiments in which we set a increasing minimum input length and concatenate consecutive turns in order to reach the threshold. Our results reveal that increasing input length yields better performance, indicating avenues for future work.
This work focuses on the research question \textit{To what extent can NLP technologies be applied to assess patients’ attachment characteristics for research and clinical applications in psychotherapy?}
\\ Our contributions are:
\begin{enumerate}
    \item The introduction of a novel method for the automatic classification of patient attachment style according to PACS \cite{Talia2017};
    \item The evaluation of our results on the only available dataset annotated with PACS;
    \item A clinical and ethical evaluation of the impact of these technologies on psychotherapy.
\end{enumerate}

\section{Related work}
\subsection{Attachment styles and their relation to psychotherapy}
An attachment constitutes an affectional bond formed between an individual and their caregiver, characterized by behaviors seeking to gain and maintain proximity to the other. This relationship and its associated behaviors bind the two individuals together in space and endures over time with the proximity-seeking behaviors being particularly prevalent during times of distress and varying between individuals in what is referred to as attachment patterns or styles \cite{Ainsworth1970, Bowlby1988}. Individuals can be classified according to four distinct attachment styles, being secure, preoccupied, avoidant and fearful (or disorganized) \cite{Mikulincer2007, Mikulincer2013} which represent patterns in behavior, thought, mental models, and emotional reactions relating to close and intimate relationships. Historically, the most reliable assessment of attachment style in adults has been conducted through the Adult Attachment Interview (AAI; \citeauthor{IJzendoorn1995}, \citeyear{IJzendoorn1995}; \citeauthor{haltigan2014adult}, \citeyear{haltigan2014adult}), however, such efforts are left to rely on the individual's subjective appraisal of their attachment characteristics \cite{Talia2017}. It was thus a paradigm-shifting development in measurement tools when the Patient Attachment Coding System (PACS) was released, given its indirect approach in which transcripts of
psychotherapy sessions are analyzed for indicators of attachment characteristics \cite{Slade2016, Talia2014, Talia2017}. The PACS enables the analysis of patient speech patterns in psychotherapy and studies how the patient manages emotional proximity with the therapist by eliciting, maintaining, or avoiding emotional attunement from the therapist. It requires a highly trained expert coder as the annotation process is deeply intricate, and it takes approximately 90 minutes to annotate a single session \cite{Talia2017}. The annotation process is conducted by assigning each sentence in the patient transcript one of 59 different discursive markers. A sentence marker could be the disclosure of a recent hurtful experience. This sentence-level annotation is then mapped to one of five categories, which could be proximity seeking, contact maintaining, exploring, avoidance or resistance. Scores on these five main scales are then mapped to three attachment categories (i.e., `secure', `preoccupied' and `avoidant'). In line with the AAI, the PACS does not include a classification for the disorganized attachment style \cite{Talia2017}. Validation studies shows very high inter-rater agreement and strong reliability and concurrent validity with the AAI \cite{Talia2017}. \\
\indent Assessment of patient attachment style is a highly valuable source of information for structuring the course of treatment, as it allows the therapist to adjust their relational style \cite{Daly2009, Slade2016, RodgersCailholBuiEtAl2010, MillerBottome2018}. This is critical for fostering a strong therapeutic alliance, vital for the outcome of therapy \cite{Baier2020}. Unfortunately, enabling such personalization is currently unfeasible given the complexity of administering the PACS, which is why an automated solution holds significant value to the delivery of psychotherapy. 

\subsection{NLP for mental healthcare}
Research investigating the application of NLP within psychotherapy has seen a considerable surge in interest after 2019, encompassing a diverse array of use-cases and applications. Three distinct clusters of research can be distilled from the current body of work.
First, speech-to-text technologies can be employed for \textit{clinical administrative work}, such as automating note-taking from psychotherapy sessions \cite{stade2024large}. These efforts comprise the most available and low-hanging fruits given the availability of the necessary models, yet they can be highly efficacious in terms of freeing up valuable time for healthcare workers hereby addressing the shortage of mental healthcare providers \cite{hoffmann2023association}.
The second cluster can be summarized under the heading of \textit{generative psychotherapy} and comprise efforts related to augmenting or even automating parts of the delivery of psychotherapy. 
Current approaches remain rudimentary, as it is observed that models tend to perform at the level expected by mediocre psychotherapist, by, for instance, being over reliant on giving advice \cite{Spiegel_Liran_Clark_Samaan_Khalil_Chernoff_Reddy_Mehra_2024}. Highly rewarding opportunities exists for hybrid modes of treatment, where, for instance, a patient who is undergoing in-person treatment could receive LLM-administered therapy at any time of their choice. One of the key advantages of introducing LLMs for psychothrapy augmentation is that the models maintain constant and ubiquitous availability, regardless of time or geography. This could theoretically allow wider access to quality health care, and address patients' need at the time of the patient's choice.
The third cluster of research relates to inferring \textit{patient, therapist and interaction specific qualities}, such as patient diagnosis, quality of therapeutic alliance or therapist adherence to treatment protocol \cite{malgaroli2023natural}. The contributions in this category hold extraordinary potential to maximize objective information gain over the course of treatment, and may ultimately revolutionize how psychotherapy is conducted \cite{stade2024large}. Inferring patient attachment style is an example of such a contribution, which, to the best of our knowledge, this study is the first to investigate.

\section{Dataset of Psychotherapy Transcripts}
In this section we describe the corpus of labeled psychotherapy transcripts, referred to as the task corpus, and the unlabeled domain data used in domain adaptive pre-training experiments.

\subsection{Task Corpus}
\label{sec:taskCorpus}
\paragraph{Source Data.}
The task data consists of 78 transcriptions, each representing a therapy session with a unique patient lasting approximately one hour.
Of these 78 transcripts, 72 sessions were transcribed for the English part of the original validation study of the PACS \cite{Talia2017}---i.e., they are double annotated with both the PACS and the AAI schema. The remaining six come from the same underling distribution, but were not included in the original study (PACS; \citeauthor{Talia2017}, \citeyear{Talia2017}) due to lack of double annotation with AAI. In our study, because we are not interested in proving the validity of PACS against AAI, as was done by \citet{Talia2017}, we use these additional six transcripts indiscriminately with respect to the original 72.
All sessions were conducted in New York, treating patients with either Brief Relational Therapy \cite{Safran2000} or Cognitive Behavioral Therapy \cite{Beck2011}.

Overall, the documents include 359,447 tokens.
Each document has been labeled with one of three classes, avoidant, secure, or preoccupied, according to its scores on the PACS scales.
These classes are represented in the dataset according to the below distribution.
\begin{itemize}
    \item {Avoidant: 20 documents (25.6\%)}
    \item {Secure: 24 documents (30.8\%)}
    \item {Preoccupied: 34 documents (43.6\%)}
\end{itemize}
The majority class in the task corpus is `preoccupied'. This is in line with the previous findings by \citet{Feeney1994} which showed that individuals with preoccupied attachment styles are more prone to seeking medical attention and more likely to report having had therapy.

\paragraph{Experimental Split.}
In our experiments, to limit the influence of therapist factors and to remain theoretically congruent with the PACS, we only afford the models access to patient speech.
Thus, for our experiments on the automatic classification of attachment style, we extract patient speech turns and label each of them according to the PACS attachment classification of its source document.
This results in 7,255 total speech turns which are individual input instances for our classification model (see Section~\ref{sec:model-setup}).
Because not all patients have the same amount of turns, this approach changes the original document-level class distribution to the following turn-level distribution:
\begin{itemize}
    \item {Avoidant: 26.7\% of speech turns}
    \item {Secure: 36.6\% of speech turns}
    \item {Preoccupied: 36.7\% of speech turns}
\end{itemize}

\begin{figure}
    \centering
    \includegraphics[width=\columnwidth]{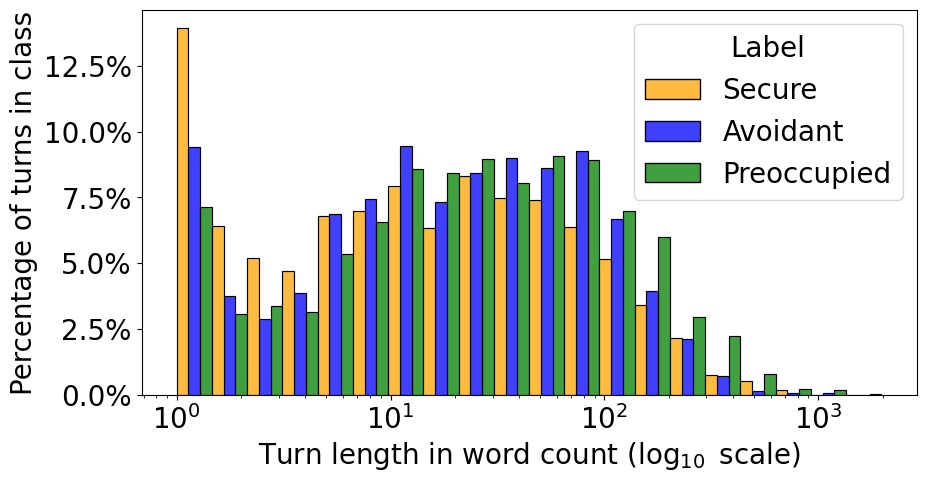}
    \caption{Distribution of patient speech turn length per class.}
    \label{fig:classBalanceTurnLength}
\end{figure}

As shown in Figure \ref{fig:classBalanceTurnLength}, classes are not equally distributed across all speech turn lengths.
Specifically, secure patients tend to have very short speech turns, while preoccupied individuals are more likely to take longer speech turns.

In our experiments, we randomly split the data by using 66 session for training and the remaining 12 for testing.
For the training, we apply 5-fold cross-validation and split the data in 85\% training and 15\% evaluation data.
The split was performed with stratification according to class balance. 
We split at the document level to avoid data leakage and ensure that the model is not learning to recognize individuals but rather generalizable speech patterns according to PACS classification.

\subsection{Unlabeled Domain Data}
\label{sec:in-domainData}
The unlabeled domain data used for domain adaptive pre-training comes from three different datasets.
First, the AnnoMI dataset \cite{Wu2022}, which consists of professionally transcribed motivational interviewing.
This dataset consists of 72,087 tokens of client speech. Second, the DAIC-WoZ dataset \cite{Gratch2014}, consisting of 275 transcripts of clinical interviews, between seven and 33 minutes in duration, conducted by an animated virtual interviewer controlled by a human interviewer in a different room. Because the annotation of this dataset can at times be ambiguous with regard to who is speaking, all utterances were included in the pre-training data. This resulted in a total of 360,488 tokens.
Finally, the HOPE dataset \cite{Malhotra2022} is made up of 212 counselling transcripts collected from various publicly available sources from which 132,942 tokens of client speech were extracted.
In total, our domain adaptive data consists of $\sim$565K tokens.

\section{Experiments}

\subsection{Model Setup}
\label{sec:model-setup}
Our classification models are based on three language encoders: The base and large versions of RoBERTa~\cite{roberta}, and MentalRoBERTa~\cite{MentalBERT}. The latter is an instance of RoBERTa base which has been further adapted on Reddit posts related to mental health.
We implement all models using the Massive Choice, Ample Tasks (MaChAmp) framework developed by \citet{MaChAmp}.
After some initial experiments, we set the learning rate to $1 \times 10^{-5}$. All the other parameters follow MaChAmp defaults.

\subsection{Domain-Adaptive Pre-Training}
To investigate whether models could learn more domain-specific representations from being exposed to more unlabeled domain data, we follow the approach proposed by \citet{Gururangan2020} and conduct domain-adaptive pre-training experiments on top of RoBERTa-base and MentalRoBERTa.
In domain-adaptive pre-training, the language model undergoes continued training on a masked language modelling (MLM) objective, using domain-relevant unlabeled data.
We set aside 20\% of domain data to validate the pre-training.
We follow MaChAmp defaults and use dynamic masking with probability 0.15 for each token for the masked language modeling.
Following recommendations from MaChAmp documentation, we duplicated the training data four times, leaving a training dataset five times the original size.
This was necessary due to the relatively low amount of data available.
Each model trained for 20 epochs, seeing each sample once, and evaluating perplexity on the development set after each epoch.
With five copies of the training data, this corresponds to training five epochs with a more frequent evaluation strategy.
The best model, as determined by the validation set performance on the MLM task, was consistently the first of 20.

\subsection{Ablation Study on Input Length}
\label{sec:input-length}
We speculate that the very short sequences (e.g, `ok', `I don't know') are less informative and therefore more challenging for the model to classify.
To further investigate whether language models would benefit from seeing longer sequences of patient speech, we conduct in-depth experiments in which we set a progressively higher minimum input length, measured in word count. To reach the threshold, we combine consecutive speech turns within the same document, and again label them with the PACS attachment classification of the source document.
These experiments were run with minimum input lengths of 50, 100, 150, and 250 words.
In these experiments, each client speech turn is included in the dataset as-is if it is at or above the defined minimum.
If it is below the threshold, it is combined with the following speech turn in the same document and then checked again.
The last speech turn in each document is included only if it meets the threshold alone or if it has already been combined with (a) previous turn(s).

\subsubsection{Ablation Study with Generative Models}
In parallel with the discriminative model setups described in Section~\ref{sec:model-setup}, we conduct some preliminary experiments testing the performance of Llama 3.1 70B~\cite{dubey2024llama3herdmodels} on classifying the attachment style from patients speech. We perform this experiments using a minimum input length of 150 words (see Section~\ref{sec:input-length}). The results obtained with Llama 3.1 were not comparable to the ones obtained with the discriminative models---accuracy of 29\%, versus the results in Table~\ref{tab:val-results} that we will present in Section~\ref{sec:validation-results}. Therefore, we focus on the discriminative approaches, leaving a deeper analysis of the generative models for future work.

\section{Results}

\subsection{Cross-Validation Results}
\label{sec:validation-results}
Validation results in the form of mean accuracy and relative standard deviation across the 5-folds are reported in Table \ref{tab:val-results}. Note that input length equal to zero corresponds to using as input each speech turn as it is without any concatenation.

Providing the model with more context, i.e., increasing the minimum input length, improves performance in most cases.
However, albeit less consistently, it also tends to increase the standard deviation, and therefore to produce more unstable results.
The increase in standard deviation is likely also attributed to the lower number of validation instances when more speech turns must be combined to reach a higher minimum length.

The domain-adapted models (i.e., MentalRoBERTa and ``+DA'') do not outperform their base counterpart.
For MentalRoBERTa the reason is likely that the domain-data on which RoBERTa was further trained (Reddit posts) is quite distant from our experimental data (therapy transcripts). Similarly, for our domain-adapted models (``+DA'') we speculate that large portion of the pre-training data is domain-relevant, but still far from the underlying distribution of our use case. Specifically, AnnoMI contains very brief sessions, and DAIC-WoZ includes virtual interviewers. 

\begin{table}
\begin{adjustbox}{width=\columnwidth}

\begin{tabular}{lccc}
\toprule
\textbf{Encoder} & \textbf{Min length} & \textbf{Accuracy} & \textbf{Std dev} \\
\midrule
\multirow{5}{*}{RoBERTa-base} & 0 & 45.28 & 7.21 \\
& 50 & 47.37 & 5.47 \\
& 100 & 51.99 & 7.42 \\
& 150 & 51.78 & 4.27 \\
& 250 & 59.86 & 9.67 \\
\midrule
\multirow{5}{*}{RoBERTa-large} & 0 & 49.27 & 3.40 \\
& 50 & 48.84 & 5.40 \\
& 100 & 54.88 & 6.94 \\
& 150 & 56.15 & 5.21 \\
& 250 & 58.67 & 11.64 \\
\midrule
\multirow{5}{*}{MentalRoBERTa} & 0 & 46.79 & 7.82 \\
& 50 & 50.65 & 6.49 \\
& 100 & 54.90 & 5.51 \\
& 150 & 53.69 & 5.31 \\
& 250 & 58.89 & 6.01 \\
\midrule
\multirow{5}{*}{RoBERTa-base +DA} & 0 & 45.17 & 6.70 \\
& 50 & 48.84 & 4.97 \\
& 100 & 49.16 & 8.66 \\
& 150 & 51.75 & 5.38 \\
& 250 & 59.40 & 7.68 \\
\midrule
\multirow{5}{*}{MentalRoBERTa +DA} & 0 & 47.30 & 5.00 \\
& 50 & 50.44 & 7.20 \\
& 100 & 54.57 & 9.59 \\
& 150 & 51.93 & 6.54 \\
& 250 & 57.22 & 9.47 \\
\bottomrule
\end{tabular}
\end{adjustbox}
\caption{Validation results. ``Min length'': Minimum input length (Section~\ref{sec:input-length}); ``+DA'': Domain Adaptation on our in-domain data (Sections \ref{sec:in-domainData})}
\label{tab:val-results}
\end{table}

\subsection{Test Set Results}
\begin{figure}
    \includegraphics[width=\columnwidth]{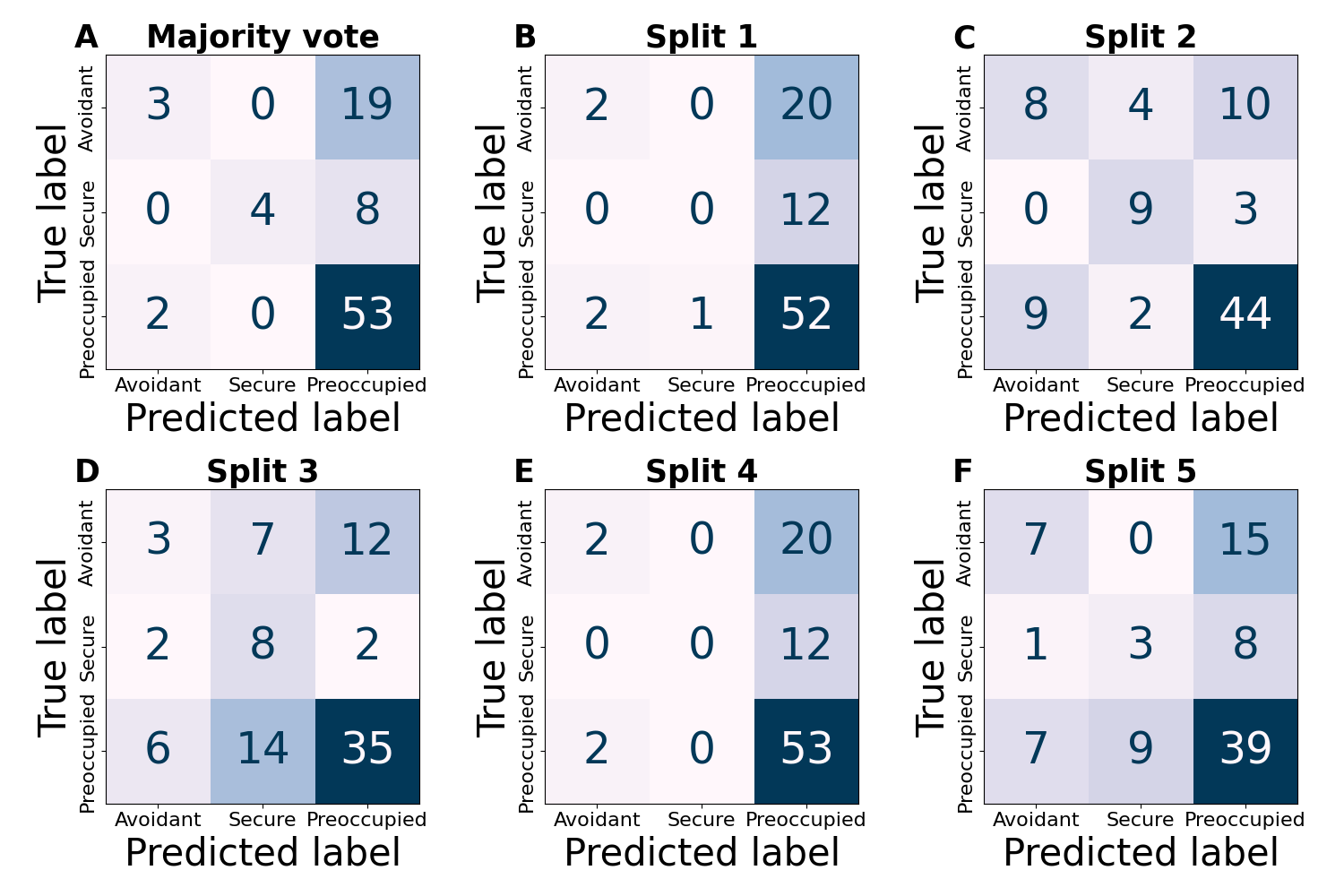}
    \caption{Test set confusion matrix for the majority vote predictions and each of five iterations of RoBERTa-large trained with minimum input length of 150 words.}
    \label{fig:ConfMatrix}
\end{figure}

Last, we test our best performing model setup on the test set (see Section~\ref{sec:taskCorpus}). 
We employ the five instances of the RoBERTa-large encoder with 150 word minimum input length. This version provides the best balance between accuracy and stability, performing among the highest accuracies with a relatively low standard deviation.
The metrics for each model are reported in Table \ref{tab:roberta-large_150 test metrics}.
The mean test set accuracy across these five models is 59.55\% (StDev=5.82).
In addition, we compute the performance of the majority voting classifier which assigns the attachment label according to the majority vote across the five models.
The accuracy score of the majority voting classifier increases to 67.42\%.
Figure \ref{fig:ConfMatrix} reports the confusion matrices of the five individual classifiers (Subplots B to F) and of the majority vote (Subplot A).
In the following section, we will further discuss these results and review implications of automating attachment style assessment in practice.

\begin{table}
    \begin{adjustbox}{width=\columnwidth}
        \begin{tabular}{lccc}
    \toprule
    Model & Accuracy & Precision & Recall \\
    \midrule
    Split 1 & 60.67 & 37.30 & 34.55 \\
    Split 2 & 68.54 & 61.42 & 63.79 \\
    Split 3 & 51.69 & 42.10 & 47.98 \\
    Split 4 & 61.80 & 37.45 & 35.15 \\ 
    Split 5 & 55.06 & 44.86 & 44.81 \\
    \midrule
    Mean & 59.55 & 44.63 & 45.26 \\
    Majority vote & 67.42 & 75.42 & 47.78 \\
    \bottomrule
\end{tabular}
    \end{adjustbox}
    \caption{Test set metrics for each instance of the RoBERTa-large model trained with minimum input length of 150 words. Precision and recall are macro-averaged.}
    \label{tab:roberta-large_150 test metrics}
\end{table}

\section{Discussion}

\subsection{Impact of Mislabeling Patients' Attachment Style}
As evident in Figure \ref{fig:ConfMatrix}, the models overpredict the `preoccupied' class. We speculate that one of the reasons contributing to this is the skewed distribution of the cross-validation set used for training (when the minimum input length is set to 150):
\begin{itemize}
    \item Avoidant: 20.70\%
    \item Secure: 31.63\%
    \item Preoccupied: 47.67\%
\end{itemize}

From a psychotherapy perspective, the error with the least impact on treatment outcome would likely be to confuse a secure patient for being either type of insecure.
Secure patients may be better able to profit from treatment regardless of the specifics of its organization \cite{MillerBottome2018, Levy2011}.
Conversely, confusing either of the insecure orientations for being secure is likely to lead to worse outcomes as therapists may alter their strategies for building the therapeutic alliance and structuring the course of treatment in counterproductive ways \cite{Daly2009}.
However, this error may be less severe for avoidant patients, as the association between this orientation and therapy outcomes is weaker \cite{Levy2011} and because this group of patients tend to hold more stable views of their therapists and working alliance \cite{Kanninen2000,Eames2000}.
Most dire are the consequences of confusing the two insecure attachment patterns.
The approaches proposed by \citet{Slade2016} and observed by \citet{Daly2009} to align best with these two attachment patterns are almost exactly mirrored, risking serious adverse outcomes from the misguided structuring of treatment according to a misclassification.

However, across the five models, the two insecure classes tend to be confused as often or more often than the secure class is confused with either of the insecure classes.
Referencing the majority vote confusion matrix (Figure \ref{fig:ConfMatrix}, Subplot A), the avoidant patients are identified with the greatest difficulty.
The `avoidant' class reaches a (low) recall of 13.64\%, indicating that the majority of avoidant patients are misclassified.
Due to the bias towards the `preoccupied' class, most mislabeled avoidant patients are instead classified as preoccupied.
If the model were applied in practice, this mislabeling would lead to the treatment of avoidant patients being systematically designed in counter-productive ways.

\subsection{Considerations on the Majority Vote Baseline}
The majority vote between all five models produces higher accuracy, precision, and recall than the majority of individual models and these scores are above the mean for each metric.
This may suggest that individual models, trained on different portions of the data, have learned to represent different aspects of the problem and that the models have not converged on similar representations.
The effect of data split indicates that these models may be able to learn more robust representations with access to more data.

The distribution between the three classes changes as more speech turns are concatenated.
The relatively small size of our dataset means that this effect is not uniform across training and test splits.
Rather, in the test split, concatenating to at least 150 words results in an increased skew and in swapping the size rankings of the `secure' and `avoidant' classes:
\begin{itemize}
    \item Avoidant: 24.72\%
    \item Secure: 13.48\%
    \item Preoccupied: 61.80\%
\end{itemize}

Most models are seemingly overconfident in predicting the `preoccupied' label, leading to a lower recall and precision as the other classes are underrepresented in their predictions.
However, most of the models do not rely only on the majority label that they ignore the `avoidant' and `secure' labels. Rather, the models attempt to classify these patients but often fails to do so correctly, reaching mean precision and recall scores of only 44.63\% and 45.26\%, respectively.
While several models do not improve in accuracy over the test set majority baseline at 61.80\% (i.e., corresponding to the percentage of `preoccupied' instances), the majority vote across models does. The models make predictions for the `secure' and `avoidant' labels rather than relying entirely on the majority `preoccupied' label (see e.g., Figure \ref{fig:ConfMatrix} Subplots C, F).
This signals that the models have learn to encode some relevant information for distinguishing the three classes and that future work could improve upon these findings to create deployment-ready models.

\subsection{Analysis on the Input Length}
\begin{figure}
    \includegraphics[width=\linewidth]{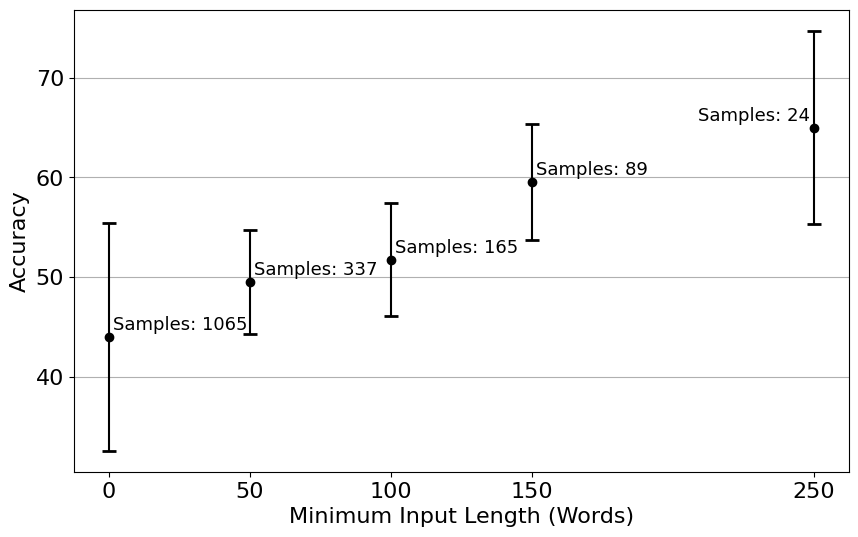}
    \caption{Mean accuracy and standard deviation in the test set over varying minimum input lengths. The error bars represent the standard deviation.}
    \label{fig: roberta-large results}
\end{figure}

To further investigate the impact of affording the model more context, Figure \ref{fig: roberta-large results} displays the mean test set accuracy and standard deviation for RoBERTa-large models across minimum input lengths.
As hypothesized, adding context to the very short speech turns tends to improve performance also in the test set.
However, it comes with a trade-off of stability as both training and testing sample sizes decrease and standard deviations in mean accuracy scores tend to increase---especially for the last data point with minimum input length of 250 words.

These results indicate that affording models more context will likely lead to improved performance. Therefore, future efforts should work towards parsing longer inputs---possibly the entire session-transcripts at once. Additionally, with longer context size, we foresee the potential of introducing therapist speech as added context.
From a technical perspective, future research could rely on language models becoming increasingly capable of dealing with longer input size. However, the greatest challenge remains the collection and annotation of psychotherapy corpora of sufficient size for language models to be trained on.

\subsection{Implications for Clinical Practice and Research}
As mentioned previously, the current process of manually annotating patients attachment using PACS is practically unfeasible, both due to the 90 minutes annotation time which exceeds the duration of the session itself, and also due to the expertise required to reliably administer the tool. Enabling the computerized inference of attachment style thus holds significant potential for increasing the quality and availability of high-impact psychotherapy treatment \cite{Slade2016}. Specifically, if the therapist could obtain an estimate of the patient's attachment style early in a treatment course, the therapist would be able to gradually adjust their expression and demeanor to match the relational issues faced by the patient \cite{Daly2009}.This has been shown to foster a stronger working alliance \cite{MillerBottome2018} and to prevent ruptures in the relationship between client and therapist \cite{RodgersCailholBuiEtAl2010}, hereby creating optimal opportunities for a favorable treatment trajectory. Some senior psychotherapists will naturally, over the course of treatment, maintain a working assumption of the attachment style of a patient, regardless of any formal assessment \cite{Slade2019, Daly2009} . However, given the intricacies related to inferring attachment and the fact that psychotherapists almost exclusively work alone, the opportunity to obtain a computerized estimate could critically function to either approve or disprove a psychotherapist's current assumption of the patient's attachment pattern. Integrating this and similar tools to current psychotherapy workflows is thus an important future consideration in the context of developing algorithmic decision-support systems \cite{binns2022human}. In sum, our approach speaks to the possibility of enabling more personalized treatment initiatives, which has so far been elusive to psychotherapy.  \\
\indent An additional benefit comes from considering the ways in which our approach can assist in the future of psychotherapy research \cite{aafjes2021scoping}. Conducting research in psychotherapy remains highly labor intensive, especially due to the need for manual transcription, and since each session lasts approximately 60 minutes data collection processes for gathering robust sample sizes are long. A consequence of this is that iteration speed in psychotherapy research is considerably lower than other branches of healthcare research. Generally, there are multiple developments in NLP technologies which may critically mitigate some of these bottlenecks, for instance, the development of automated speech recognition software. However, the most significant implications lies within enabling researcher access to information which would otherwise be unfeasible to gather, such as attachment style. Our work explores the possibility of automatically inferring patient attachment style, hereby giving diverse researchers the opportunity to control for a broader range of factors and constructs when, for instance, evaluating the effect of a randomized controlled trial. This holds the opportunity to strengthen the quality of psychotherapy research and increase the speed with which new findings and treatments may be available for patients. \\
\indent Amidst their great promise, it remains vital to consider potential adverse effects and ethical implications of these developments.
In the case of using language models in psychological research and practice, \citet{Demszky2023} point to two primary sources of harm arising from bias. First, \textit{representational harms} can arise when some socio-demographic groups are represented in unfavorable ways by the language models. Second, \textit{allocational harms} may arise when algorithms are employed in distributing resources (e.g. loans, access to therapy) but do so differentially according to biases present in the training data. Cross-domain analyses reveal that the construct of attachment remains strikingly similar across gender, language, and culture \cite{Bakermanskranenburg2009}, rendering automatic attachment inference more scalable than other psychological constructs. These findings by no means omit the need for careful adaptation studies, where especially cross-cultural expression patterns should be analyzed in detail. Finally, as the proposed model relies on psychotherapy \textit{transcripts}, developments in automatic speech recognition technologies for under-resourced languages currently serve as a short-term bottleneck for enabling more widespread adoption. 

\section{Conclusion}
The automation of attachment assessment in psychotherapy offers significant potential for both research and clinical practice. Our study represents a preliminary exploration of this possibility using limited data. Our results indicate that developments in NLP methodologies may become effective in distinguishing patient attachment patterns, although more research and larger datasets are needed.

Compared to manual approaches like the PACS, automated assessment can provide repeated measurements throughout treatment and scale more effectively for research. This enables a deeper understanding of therapeutic processes and mechanisms, as well as more targeted and personalized interventions.
PACS annotators require extensive training (approximately 30 hours) and spend around 90 minutes analyzing each 60-minute session.
For a therapist seeing five clients per day, the weekly time savings from automating annotation can exceed 30 hours.

Considering the lack of studies exploring the potential of NLP technologies in the automated assessment of attachment style, we suggest that future research in this field would focus on data collection and annotation, particularly to enable research with longer inputs.
Additionally, investigating the linguistic features that the NLP models rely on would shed light on the underlying mechanisms of talk therapy. This is essential for transparency and interpretability, and has the potential to inform the development of more effective interventions, based on psychotherapy theories.

\section*{Limitations}
A limitation of our exploratory work is the small size of the dataset. As mentioned, the time-consuming nature of administering the PACS also means the general availability of correctly annotated transcripts are scarce. Further, as a client during psychotherapy reveals the most intimate and private aspects of their life, transcripts are also governed by high privacy concerns, placing an additional bottleneck on available documents. 
On the technical side, our model architecture is constrained by the limited computational resources of our institutions. 
Specifically, the data was stored in the computational system of a humanistic department, and given the privacy concerns just mentioned, could not be moved to a higher performing cluster.

\section*{Acknowledgments}
We thank the NLPnorth group at ITU for feedback on earlier version of this paper. 
EB is supported by a research grant (VIL59826) from VILLUM FONDEN. MLT is supported by a research grant from the Pioneer Centre for AI, DNRF grant number P1. 

\bibliography{anthology,custom}

\end{document}